\documentclass{article}
\usepackage{spconf,amsmath,graphicx}
\usepackage{makecell} % for ablation table
\usepackage{booktabs}  % optional, for nicer tables
\usepackage{enumitem}
\setlist{nosep, leftmargin=14pt}
\usepackage{adjustbox}
\usepackage[numbers]{natbib}
\usepackage{float}
\usepackage{mwe} % to get dummy images
\usepackage{array, multirow, rotating, booktabs, caption, adjustbox}
\title{TractRLFusion: A GPT-Based Multi-Critic Policy Fusion Framework for Fiber Tractography}

\name{
\begin{tabular}{c}
Ankita Joshi$^{1}$, Ashutosh Sharma$^{1}$, Anoushkrit Goel$^{1}$, Ranjeet Ranjan Jha$^{2}$,\\ Chirag Ahuja$^{3}$, Arnav Bhavsar$^{1}$, Aditya Nigam$^{1}$
\end{tabular}
}
\address{
    $^1$ IIT Mandi, $^2$ IIT Patna, $^3$ PGIMER Chandigarh
}

\begin{document}
\maketitle
\begin{abstract}

% Tractography is a fundamental task in neuroimaging, enabling noninvasive reconstruction of white matter fiber pathways in the brain. It plays a vital role in understanding structural brain connectivity and guiding neurosurgical planning. Early tractography methods have primarily relied on classical deterministic and probabilistic algorithms, while recent advances have explored supervised deep learning (DL) and deep reinforcement learning (DRL) techniques.
% A persistent challenge in tractography is accurately reconstructing white matter tracts while minimizing anatomically implausible or spurious connections. Many algorithms inherently face this trade-off, which limits the reconstruction accuracy. To address this, we propose TractRLFusion, a novel GPT-based policy fusion framework that integrates multiple RL policies through a data-driven fusion strategy. The method includes a two-stage training data selection process to enable effective fusion, followed by a multi-critic policy fine-tuning phase. We demonstrate that TractRLFusion outperforms individual RL policies as well as state-of-the-art classical and DRL-based methods across three benchmark datasets: HCP, ISMRM, and TractoInferno.
Tractography plays a pivotal role in the non-invasive reconstruction of white matter fiber pathways, providing vital information on brain connectivity and supporting precise neurosurgical planning. Although traditional methods relied mainly on classical deterministic and probabilistic approaches, recent progress has benefited from supervised deep learning (DL) and deep reinforcement learning (DRL) to improve tract reconstruction.
A persistent challenge in tractography is accurately reconstructing white matter tracts while minimizing spurious connections. To address this, we propose TractRLFusion, a novel GPT-based policy fusion framework that integrates multiple RL policies through a data-driven fusion strategy. Our method employs a two-stage training data selection process for effective policy fusion, followed by a multi-critic fine-tuning phase to enhance robustness and generalization. Experiments on HCP, ISMRM, and TractoInferno datasets demonstrate that TractRLFusion outperforms individual RL policies as well as state-of-the-art classical and DRL methods in accuracy and anatomical reliability.
\end{abstract}
\vspace{-0.25cm}
\section{Introduction}
\label{sec:intro}
\vspace{-0.2cm}
White matter fiber tractography\cite{basser2000vivo} is a key technique to study brain structural connectivity, neurological disorders, and to support precise neurosurgical planning \cite{essayed2017white}. It enables in vivo and non-invasive mapping of white matter pathways using diffusion MRI (dMRI) data. dMRI captures the anisotropic diffusion (movement) of water molecules within white matter tissue, as water diffuses more readily along the axonal fibers. This directional dependence is typically modeled using fiber orientation distribution functions (fODFs), allowing inference of white matter pathways.
Traditional tractography algorithms are generally categorized as \textit{deterministic}, \textit{probabilistic}, and \textit{global} approaches. \textit{Deterministic methods}\cite{basser1998fiber,cheng2014tractography} reconstruct streamlines by following the principal diffusion direction at each step, but often fail in regions with crossing, branching, or other complex fiber configurations.
In contrast, \textit{probabilistic methods}\cite{berman2008probabilistic} estimate a distribution of possible pathways at each step, improving coverage, but at the cost of increased false positives (FPs). Global methods\cite{reisert2011global} attempt to handle the FP-FN trade-off in tractography\cite{schilling2019challenges,kamagata2024advancements}, improving overall consistency but at the cost of high computational cost and limited generalizability.
Supervised Learning based methods rely on ground-truth tracts for training. For example, a Random Forest Classifier\cite{neher2015machine} predicts the next streamline direction from a set of candidate directions and past estimated steps.
Similarly, DL methods such as Learn-to-Track\cite{poulin2017learn} and DeepTract\cite{benou2019deeptract} formulate tractography as a sequential decision making task, employing GRUs to model streamline paths and probabilistic fiber orientation distributions, respectively.
In parallel, DRL methods\cite{theberge2021track,theberge2024matters,theberge2024tractoracle,joshi2024tract} learn tractography policies through exploration rather than direct supervision.

Interestingly, we observe a similar FP-FN trade-off among both deterministic and stochastic RL policies. This observation motivates our investigation of the fusion of RL policies to improve tractography performance, a strategy that has proven to be effective in other domains, such as robotics and AI games to improve generalization and robustness\cite{lillicrap2015continuous,marinov2024offline}. Traditional ensemble strategies for RL policy, including decision-level aggregation (e.g., voting or averaging), often fail to capture semantic context or state history. Conversely, more complex ensemble methods, such as Ensemble Policy Gradient or hierarchical ensembles, introduce significant system complexity due to agent synchronization, parameter sharing, and joint optimization requirements, making them difficult to implement, debug, and maintain.

\begin{figure}[t]
    \centering

    \includegraphics[width=\linewidth, height=6cm]{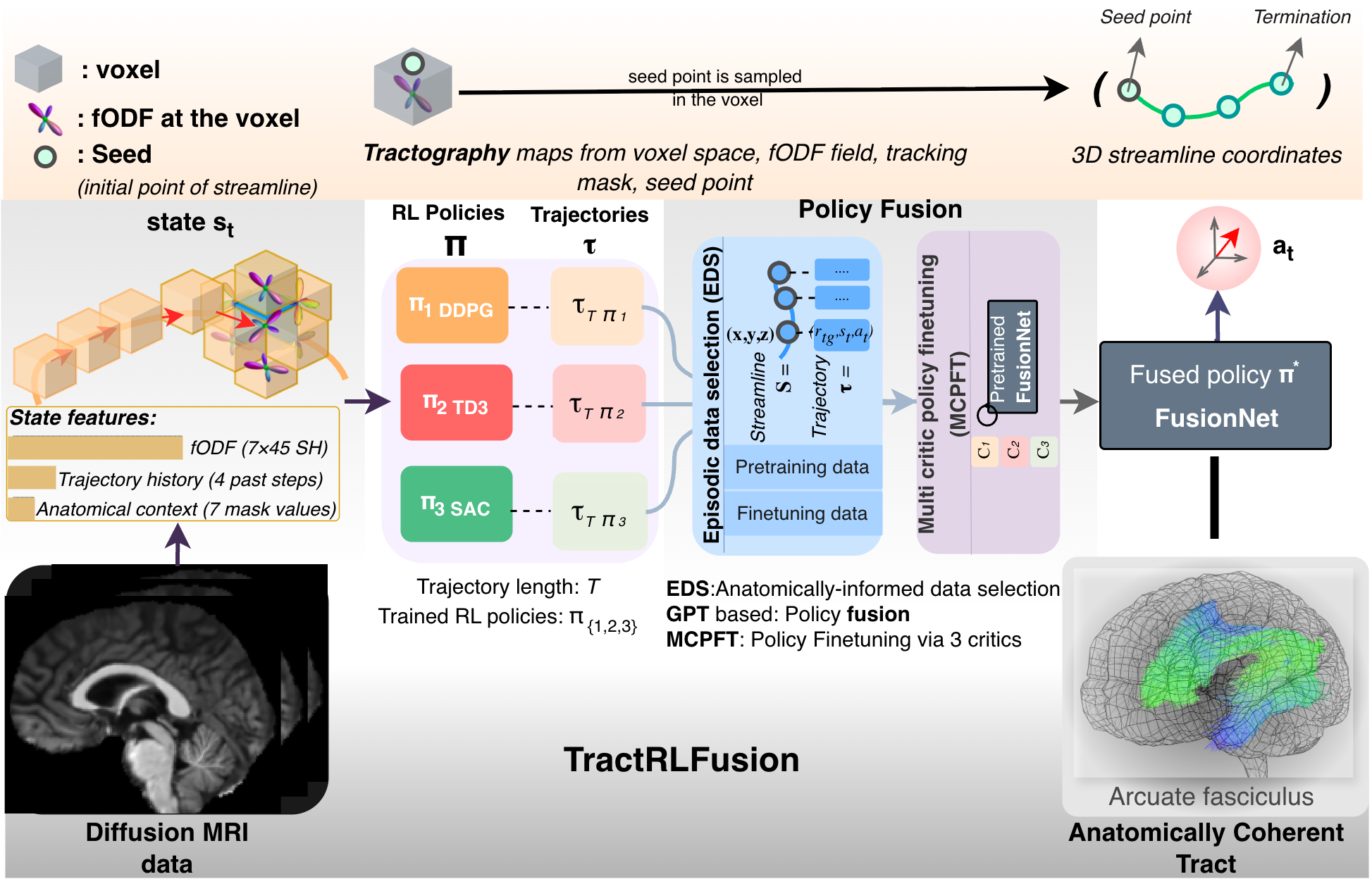}
    \caption{\textbf{TractRLFusion}: An RL policy fusion framework for tractography, comprising EDS, FusionNet and MCPFT.}
    % A policy-fusion framework consisting of EDS-based trajectory selection, FusionNet, and MCPFT}
    % multi-critic fine-tuning to generate anatomically coherent tracts}
    % \caption{\textbf{TractRLFusion}: A policy-fusion framework constituting EDS-based trajectory selection, FusionNet, and multi-critic policy fine-tuning to generate anatomically coherent tracts}
    % \caption{\textbf{TractRLFusion}: Fiber tractography environment is modeled as a Partially Observable Markov Decision Process (POMDP), where the agent only receives a local observation $o_t$. A set of Markovian policies $\Pi = \{\pi_k\}_{k=1}^3$ interact with the environment, generating interaction histories $h_t$. Theoretically, this forms a belief state $b(s_t)$ about the true, unobservable state. Our \textbf{Policy Fusion} block, comprising Episodic Data Selection and Multi-Critic Policy Fine-tuning, processes these histories to learn a single Autoregressive Fused Policy ($\pi_\phi^*$). The policy predicts actions conditioned on history, enabling the generation of anatomically coherent tracts.}
    \label{fig:pipeline}
    \vspace{-0.6cm}
\end{figure}

To address these challenges, we propose \textbf{TractRLFusion} (Fig. \ref{fig:pipeline}), a data-driven fusion framework to combine multiple RL policies in tractography. In this work:
\begin{itemize}
    \item We introduce \textbf{TractRLFusion}, a novel GPT-based policy fusion framework that integrates \textbf{Episodic Data Selection (EDS)} to select tract-specific trajectories with anatomical precision, enabling data-driven fusion.
    \item We propose \textbf{Multi-Critic Policy Fine-Tuning (MCPFT)} to enhance policy fusion, achieving \textbf{robust}, \textbf{scalable}, and \textbf{generalizable} tractography across diverse datasets.
    \item Extensive \textit{comparative} evaluation against state-of-the-art tractography methods demonstrate its \textbf{superior performance, effectiveness}, and generalizability.
\end{itemize}
% \vspace{-0.3cm}
% To the best of our knowledge, this is the first policy fusion framework proposed for tractography.
% \textbf{The code will be released on GitHub.}

To our knowledge, this is the first policy-fusion framework for tractography. \textbf{Code will be released on GitHub.}

\vspace{-0.4cm}
\section{Methodology}
\vspace{-0.3cm}
We fuse three independently trained policies, namely TD3 \cite{fujimoto2018addressing}, SAC \cite{haarnoja2018soft}, and DDPG \cite{lillicrap2015continuous}, to combine their complementary strengths in tractography. While SAC learns in an exploration-driven manner, resulting in higher tract coverage but increased false positives, TD3 and DDPG exhibit more conservative tracking behavior, yielding lower coverage with fewer false positives, as shown in Table \ref{table:exp:combined}. The TractRLFusion framework comprises three key components: Episodic Data Selection (\textbf{EDS}), \textbf{FusionNet}, and Multi-Critic Policy Finetuning (\textbf{MCPFT}). 
To enable data-driven fusion, we sample trajectories from trained RL policies, where a trajectory represents a sequences of state, action, and return-to-go tuples obtained from an RL policy. A trajectory of length T is defined as: $\tau = (s_{0}, a_{0}, r_{tg_{0}}, s_{1},.... s_{T}, a_{T}, r_{tg_{T}})$.

\textbf{(i)} EDS samples a mix of trajectory data from each of the three policies to train FusionNet.
\textbf{(ii)} FusionNet is then trained in two stages on EDS data to learn the fusion policy.
\textbf{(iii)} Finally, the FusionNet policy is refined using MCPFT to enhance stability and generalization.

FusionNet learns a tract-specific policy for each targeted white matter bundle. To this end, we train and evaluate our policies within tract-specific masks from \cite{joshi2024tract}.
Among the major tracts, we focus on the occipital and pre/post central gyri regions of the Corpus Callosum (CC), the left and right Corticospinal Tract (CST), the Arcuate Fasciculus (AF), and the Cingulum (CG), following previous work\cite{theberge2024matters}. 

% \textbf{Tract-Specific RL policy training:} With an environment similar to \cite{theberge2021track}, training is conducted via exploration within tract-specific masks of \textbf{5 train subjects} from TractoInferno dataset\cite{poulin2022tractoinferno} (Table \ref{tab:hyperparam}). 

\textbf{Tract-Specific RL policy Training:} Training is conducted through exploration in an environment similar to \cite{theberge2021track}, but constrained within tract-specific masks of \textbf{five training subjects} from the TractoInferno dataset\cite{poulin2022tractoinferno} (Table \ref{tab:hyperparam}).
The reward is defined as the product of the action vector's ($\vec{a}_t$) alignment with the most aligned fODF peak ($p$) and its dot product with the previous tracking direction ($\vec{u}_{t-1}$):
\vspace{-0.2cm}
\begin{equation}
    r_t = \left| \max_{\vec{p}_i} \left( \vec{p}_i \cdot \vec{a}_t \right)\right| \times \left( \vec{a}_t \cdot \vec{u}_{t-1} \right)
    \label{eqn:reward}
\end{equation}
% \vspace{-0.2cm}
\textbf{Episode termination conditions} are \textbf{(i)} exceeding 530 steps ($\equiv$200 mm fiber), \textbf{(ii)} leaving the tract mask, or \textbf{(iii)} an angle $>$ 60$\textdegree$ with the previous segment.
Subsequent sections describe each component of Tract-RLFusion in detail.
% \textbf{Episode termination conditions} include \textbf{(i)} exceeding the maximum episode length of 530 (corresponding to max. fiber length=200mm), \textbf{(ii)} exiting tract mask, and \textbf{(iii)} angle $>$ 60 degrees with the previous fiber segment.
% The following paragraphs explain each component of Tract-RLFusion.

\vspace{-0.4cm}
\subsection{Episodic Data Selection}
\vspace{-0.1cm}
It is the data preparation process for the pre-training and fine-tuning stages of FusionNet model. It involves following steps:
% to obtain "expert" demonstrations from three actor-critic RL policies (TD3, DDPG and SAC).

\textbf{(i)} Tracking is performed using each trained policy (TD3, SAC, and DDPG) from a common batch of neighboring seed points sampled from the tract mask of a given subject. This produces streamlines \{(x,y,z)\} along with their corresponding trajectory tuples {$(s,a,r_{tg})$}.
% (see Fig. \ref{fig:EDS}(i)).

\textbf{(ii)} Trajectories containing fewer than 47 transitions (equivalent to approximately $20$mm for a 1mm$^{3}$ voxel size and 0.375 step size) are filtered-out. The remaining tracjectories then undergo batch-wise \textbf{across-policy} and \textbf{within-policy} selection for subsequent processing.

% Per subject.
% \textbf{(ii)} Trajectories shorter than $47$ transitions (or $20mm$ fiber length for voxel size of $1mm^{3}$ isometric and step size of 0.375) are filtered-out and the rest undergo batch-wise \textbf{across-policy} selection and \textbf{within-policy} selection.

\textbf{(iii)} \textbf{Within-policy selection}: Fifteen reference streamlines are selected from the atlas tract\cite{yeh2018population} using the farthest streamline sampling. For each policy, we filter streamlines to retain only the trajectories that lie within a 5mm MDF (mean direct flip) distance \cite{li2023farthest} from the reference streamlines, thus selecting the most anatomically accurate trajectories.

% \textbf{(iii)} \textbf{within-policy selection}: 15 reference streamlines are selected from the atlas \cite{yeh2018population} tract using farthest streamline sampling. Streamlines for each policy are filtered such that trajectories of streamlines lying within 5mm MDF (mean direct flip) distance \cite{li2023farthest} of the reference streamlines are selected, thereby selecting the most anatomically accurate trajectories.

% enabling selection of the best trajectories in terms of anatomical shape of any given tract.

% \textbf{(iii)} 15 reference streamlines are selected from the atlas \cite{yeh2018population} tract using farthest streamline sampling \cite{li2023farthest}. Streamlines for each policy are filtered such that trajectories of streamlines lying within 5mm MDF (mean direct flip) distance \cite{li2023farthest} of the reference streamlines are selected. This \textbf{within-policy selection} enables the selection of the best trajectories in terms of anatomical shape of any given tract.

% \textbf{(iii)} 15 reference streamlines are selected from the atlas tract using farthest streamline sampling \cite{li2023farthest}. Streamlines for each policy are filtered such that trajectories of streamlines having MDF (mean direct flip) distance \cite{li2023farthest} $<$ 5mm with the reference streamlines are selected. This \textbf{within-policy selection} enables the selection of the best trajectories in terms of anatomical shape of any given tract.

\textbf{(iv)} \textbf{Across policy selection} selects one policy out of the three (TD3, SAC, DDPG), based on the maximum expected Q-value (normalised to incorporate different scales) of its MDF-filtered trajectories. These trajectories are selected for the finetuning dataset.  In contrast, for the pretraining dataset, Q-value-based selection is performed directly on the unfiltered trajectories.
% Whereas, for pretraining dataset, Q-value based selection happens directly on unfiltered trajectories.
% , as depicted in Fig. \ref{fig:EDS}. 
% The seeds in a batch are neighbouring seeds in/of that tract, thereby dividing it into smaller regions for selecting best policy in terms of fiber anatomy(shape) for the/any given tract. This is repeated for all batches of seeds which vary in size for different WM tracts. 

Neighboring seed points are selected in each batch, enabling EDS to select the best trajectories from all regions of the given tract. This process is repeated across all tracts of the five training subjects, resulting in approximately 150000 pretraining and 50000 finetuning trajectories after length-based and random selection. In particular, the pretraining dataset comprises trajectories from all tracts, whereas the \textbf{finetuning data consists of tract-specific trajectories}.
% \textit{\textbf{Note:}} Within-policy trajectory selection precedes Across-policy selection because we want to first extract the best trajectories relevant to tracking process for a region (introducing shape similarity/ info) and then only decide , which one to select for that region (set of neighbouring seeds) using Q-value comparison.
% OR
% Note: Within-policy trajectory selection precedes across-policy selection to first extract the best region-specific trajectories (e.g., via shape similarity), before using Q-value comparison to choose  final trajectories for finetuning.

\vspace{-0.4cm}
\subsection{FusionNet (Architecture and Training)}
\vspace{-0.1cm}
Central to our framework, FusionNet is a 4-layer GPT \cite{radford2018improving} (one attention head, 128-dim embeddings, context length 40, ReLU activation, and dropout rate of 0.1) that models RL trajectories as sequences. It undergoes general pretraining and tract-specific fine-tuning on expert trajectories from EDS. Specifically, the first three layers are trained on mixed trajectories from the EDS pretraining data over 30 iterations, while the final layer is fine-tuned for 10 iterations using tract-specific trajectories from the EDS finetuning data. This two-stage training strategy (Table \ref{tab:fusionnet_params}) is designed to first capture broad trajectory patterns and subsequently adapt to the fine-grained tract-specific details. We employ a five-step angular loss ($\mathcal{L}_{dist_{cos}}$) between predicted ($\hat{\textbf{a}}$) and actual ($\textbf{a}$) actions:
% , defined as:
% {\setlength{\abovedisplayskip}{5pt}\setlength{\belowdisplayskip}{5pt}
\vspace{-0.3cm}
\begin{equation} \label{supervised_loss}
    \mathcal{L}_{dist_{cos}} = \sum_{t=2}^{l_{context}-2} \left( \sum_{i=-2}^{2} \cos^{-1} \left( \textbf{a}_{t+i} \cdot \hat{\textbf{a}}_{t+i} \right) \right)
% \vspace{-0.2cm}
\end{equation}

% This 2-stage training is devised to capture broad patterns as well as downstream finetuning to incorporate intricate details of a tract.
% \noindent 
The model was trained with AdamW optimizer, hyperparameters tuned to maximize Accumulated sum of rewards and Dice Scores. Additional training with supervised loss (Eq. \ref{supervised_loss}) led to performance saturation, showing limited improvement.
% We observed that a

% Additional details regarding hyperparameter tuning can be found in the Appendix. 

 % The first 3 layers are trained on mixed trajectories from EDS pretraining data for 30 iterations ($10,000$ updates per iteration, batch size=128). The last layer is finetuned for 10 iterations on tract-specific trajectories from EDS-Finetune data.

% Model training was performed for each tract, using AdamW optimizer with a learning rate of 1e-4 and a weight decay of 1e-4. A warmup phase of 10000 steps was applied. ReLU activation function was used with a dropout of 0.1. 
% Further training with the supervised loss (ref. eq. \ref{supervised_loss}) saturated performance, shows limited improvement.

\vspace{-0.35cm}
\subsection{Multi-critic Policy Fine-tuning (MCPFT)}

We fine-tune FusionNet using MCPFT in an actor-critic setup with critic networks derived from SAC, DDPG, and TD3. 
\vspace{-0.6cm}
% \begin{equation} 
% \label{eq:actorloss}
% L_{\text{actor}} = L_{dist_{cos}} + \sum_{k=1}^K L_{\text{critic}, k},
% \end{equation} 
% with $L_{\text{critic}, k}$ over a trajectory of length $l_{context}$ is, \\
% \vspace{-0.2cm}
% \begin{equation}
% \label{eq:criticloss}
% L_{\text{critic}, k} = -\sum_{t=0}^{l_{context}} Q_{(\pi_k)}^C(s_t, a_t).
% \end{equation}

\begin{equation}
\label{eq:actorloss}
\mathcal{L}_{\text{actor}} =
\mathcal{L}_{\text{dist}_{\text{cos}}} +
\sum_{k=1}^{K=3}\!
\underbrace{\left[-\!\sum_{t=0}^{l_{\text{context}}}
Q^{C}_{(\pi_k)}(s_t,\hat{a}_t)\right]}_{\mathcal{L}_{\text{critic},k}}
\end{equation}

\vspace{-0.3cm}
The actor (\textbf{FusionNet}) is optimized using the composite loss $L_{actor}$ (Eq. \ref{eq:actorloss}), which combines the supervised loss $L_{dist_{cos}}$ with the aggregated loss from our \textit{K = 3} critics, where each $\mathcal{L}_{critic,k}$ is formulated to maximize the critics' value estimates (Eq. \ref{eq:actorloss}). 
Standalone aggregated critic loss ($\sum_{k=1}^{3} \mathcal{L}_{\text{critic}, k}$) caused a performance drop after a few updates, followed by saturation below the prior levels. Our observations during training indicate the ability of $\mathcal{L}_{dist_{cos}}$ to preserve prior actor knowledge, while the critic term further refines its policy. We observed that delayed critic updates prevent saturation from unstable value estimates. Training parameters are summarized in Table \ref{tab:fusionnet_params}.

\vspace{-0.3cm}
\subsection{Tracking Process}
% \vspace{-0.4cm}
% \noindent \textbf{2.4  Tracking Process}\\
% \noindent 
\vspace{-0.1cm}
Using the FusionNet policy, tracking is performed for different tracts within their respective tracking masks, by seeding at 7 seeds per voxel. Return-to-go ($r_{tg0}$) is initialized to 300 for FusionNet, and dataset-specific step sizes (0.375mm for TractoInferno, 0.468mm for HCP, 0.75mm for ISMRM) are set. As a postprocessing step, tract filtering is performed using Fast Streamline Search \cite{st2022fast} with atlas reference tracts \cite{francois_rheault_2023_7950602}. 

\begin{table}[hbt!]
\centering
\fontsize{9}{9.5}\selectfont
\vspace{-0.2cm}
\begin{tabular}{p{2.2cm}|p{5.5cm}}
\hline
\textbf{Category} & \textbf{Details} \\
\hline
\textbf{State} \newline \textbf{Features} & 
- 45 SH \textit{coefficients} (6 neighbors+1) \newline
- Last 4 past directions (12 values) \newline
- Tract mask value (6 neighbors+1)\newline
% - 6 neighbors + 1 binary tract-specific mask = 7 \textit{voxels}\newline
% - Total Observation dimension: ((45+1)$\times$ 7 ) + 12 = 334 \\
% - Total: ((45+1)$\times$ 7 ) + 12 = 334 dimensions\\
- Concatenated to: 334 dimensions\\
\hline
\textbf{Architectures} \newline (Actor, Critics) & 
% - \textbf{TD3}/\textbf{SAC}/\textbf{DDPG} (Actor, Critics): \newline 
- 3-layer ReLU-activated FCNNs \newline
- Neurons per hidden layer: 1024 \\
\hline
\textbf{Train Data} \cite{poulin2022tractoinferno}& 
% - Train Subjects: 5 (IDs: 1030, 1079, 1119, 1180, 1198) \cite{poulin2022tractoinferno}\\
- IDs:1030, 1079, 1119, 1180, 1198\\

% \textbf{Train Subjects}& 
% % - Train Subjects: 5 (IDs: 1030, 1079, 1119, 1180, 1198) \cite{poulin2022tractoinferno}\\
% - 1030, 1079, 1119, 1180, 1198 \cite{poulin2022tractoinferno}\\

\hline
\textbf{Hyperparameters} & 
- Batches: 50 per subject \newline
- Batch size: 4096 transitions \newline
- Seeding: 7/voxel; Step size: 0.375 \\
\hline
\textbf{Policy} \newline \textbf{Hyperparameters} & 
- \textbf{TD3}: $\eta$ = 8.56e-6, $\sigma$ = 0.334, $\gamma$ = 0.776 \newline
- \textbf{SAC}: $\eta$ = 3.7e-5, $\sigma$=0.4, $\gamma$=0.89, $\alpha$=0.076 \newline
- \textbf{DDPG}: $\eta$ = 8.56e-6, $\sigma$ = 0.35, $\gamma$ = 0.5 \\
% \hline
\end{tabular}
\vspace{-0.2cm}
\caption{Summary of Features, Architecture, Train Subjects, and Hyperparameters for training Tract-specific RL policies.}
\label{tab:hyperparam}
\vspace{-0.3cm}
\end{table}

\vspace{-0.3cm}

\begin{table}[hbt!]
\centering
\fontsize{9}{9.5}\selectfont
% \begin{tabular}{l|p{5.2cm}} % Using p{} column for vertical stacking
\begin{tabular}{p{2.2cm}|p{5.5cm}}
\hline
\textbf{Category} & \textbf{Details} \\
\hline
% \textbf{Architecture} & 
% - GPT Layers: 4  \newline
% - Embedding Dim: 128 \newline
% - Attention Heads: 1 \newline
% - Context Length ($l_{context}$): 40 \newline
% - Activation: ReLU; Dropout: 0.1 \\
% \hline
\textbf{FusionNet} Training & 
- Optimizer: AdamW ($\eta$: 1e-4) \newline
% - Weight-Decay: 1e-4 \newline
- Batch Size: 128; Updates/iter: 1e4 \newline
- Iters:30(General),10(Tract-specific) \newline
- Warmup-Steps: 1e4\\
\hline
\textbf{MCPFT}-based \newline Finetuning & 
- $\eta$: 1e-4 ; Batch Size: 512 ; Iters: 25 \newline
- Updates/iter: 1000 (actor), 1 (critic) \\
\hline
\end{tabular}
\vspace{-0.3cm}
\caption{Summary of FusionNet training parameters.}
\label{tab:fusionnet_params}
\vspace{-0.4cm}
\end{table}

\vspace{-0.2cm}
\section{Experiments and Results}
\vspace{-0.3cm}
\begin{table}[hbt!]
% [htbp]
\centering
\fontsize{9}{9.5}\selectfont
\renewcommand{\arraystretch}{0.85}
% \vspace{-0.5cm}
% ============ (a) HCP + TractoInferno ============
\textbf{(a) CG and CST/PYT}
\begin{tabular}{ |p{0.44cm}|p{1.3cm}|p{0.5cm}p{0.5cm}p{0.5cm}|p{0.5cm}p{0.5cm}p{0.5cm}| }

 \hline
  &  & \multicolumn{3}{c|}{\textbf{CG}} & \multicolumn{3}{c|}{\textbf{CST/PYT}} \\ \cline{3-8}
\textbf{Data} & \textbf{Algorithm} & \textbf{\textbf{Dice}\textuparrow} & \textbf{OL}\textuparrow & \textbf{OR}\textdownarrow & \textbf{Dice}\textuparrow & \textbf{OL}\textuparrow & \textbf{OR}\textdownarrow \\ \cline{1-8}
 % & TractSeg &  &  &  &  &  &  &  &  &  &  &  &  \\
 & DET & 53.9  & 44.1 & 18.9 & 62.9 & 51.8 & 11.7 \\
 % & PROB & \tikzmark{CG hcp 1 top}56.9 & 48.7 & 22.3\tikzmark{CG hcp 1 bottom} & \tikzmark{CST hcp 1 top}74.7 & 71.8 & 20.0\tikzmark{CST hcp 1 bottom} & \tikzmark{AF hcp 1 top}56.5 & 45.8 & 13.5\tikzmark{AF hcp 1 bottom} & \tikzmark{CC hcp 1 top}74.9 & 84.1 & 41.6\tikzmark{CC hcp 1 bottom} \\ \cline{2-14}
 & SAC & 56.2 & 47.3 & 20.2 & 67.7 & 60.8 & 17.4 \\
 & DDPG & 48.0 & 35.8 & 12.8 & 66.1 & 60.4 & 20.4 \\
% \parbox[c][0pt][c]{\linewidth}{\centering\rotatebox[origin=c]{90}{HCP \shortcite{wasserthal2018tractseg}}}

\parbox[c][0pt][c]{\linewidth}{
  \centering
  \rotatebox[origin=c]{90}{
    \parbox{3cm}{\centering HCP}
  }
} & TD3 & 49.2 & 37.5 & 14.7 & 55.7 & 46.2 & 15.8 \\
 & TRLF$_{S}$ & 54.0 & 41.6 & 12.3 & 61.6 & 54.1 & 15.7 \\
  & $\pi_{maxQ}$ & 56.3 & 47.5 & 20.7 & 67.9 & 59.5 & 14.9 \\
  & $\pi_{avg}$ & 56.0 & 46.6 & 19.6 & 69.2 & 63.7 & 17.6 \\
 %mod----
 & \textbf{FusionNet} & \textbf{56.6} & 47.2 & 19.2 & \textbf{69.4} & 62.5 & 17.0 \\ \cline{2-8}
 & TractSeg & 76.8 & 73.1 & 17.2 & 80.8 & 73.2 & 7.8 \\
 & \textbf{FusionNet*} & \textbf{77.0} & 86.7 & 38.2 & \textbf{82.2} & 79.8 & 14.4 \\
  % & SAC & 56.2 & 47.3 & 20.2 & 67.7 & 60.8 & 17.4 & 54.6 & 42.4 & 11.4 & 70.4 & 80.4 & 48.1 \\
\Xhline{2\arrayrulewidth} % Thicker line
 & DET & 59.7 & 56.8 & 33.0 & 71.5 & 77.8 & 39.8 \\
 % & PROB & \tikzmark{CG ttoi 1 top}66.3 & 66.9 & 35.2\tikzmark{CG ttoi 1 bottom} & \tikzmark{CST ttoi 1 top}76.2 & 76.0 & 23.3\tikzmark{CST ttoi 1 bottom} & \tikzmark{AF ttoi 1 top}56.3 & 52.8 & 39.1\tikzmark{AF ttoi 1 bottom} & \tikzmark{CC ttoi 1 top}59.7 & 53.2 & 25.9\tikzmark{CC ttoi 1 bottom} \\ \cline{2-14}
 & SAC & 63.1 & 70.9 & 55.9 & 73.0 & 70.3 & 22.0 \\
 & DDPG & 59.9 & 55.7 & 30.4 & 69.2 & 60.7 & 14.3 \\
\parbox[c][0pt][c]{\linewidth}{
  \centering
  \rotatebox[origin=c]{90}{
    \parbox{3cm}{\centering TractoInferno}
  }
} & TD3 & 57.5 & 52.4 & 29.9 & 69.6 & 62.1 & 15.9 \\
& TRLF$_{S}$ & 62.5 & 63.3 & 39.8 & 69.8 & 64.1 & 19.1 \\
 & $\pi_{maxQ}$ & 63.5 & 71.5 & 53.9 & 73.9 & 70.6 & 21.1 \\
  & $\pi_{avg}$ & 62.3 & 71.2 & 60.9 & 73.2 & 71.4 & 23.5 \\
 %mod----
 & \textbf{FusionNet} & \textbf{64.0} & 65.9 & 41.6 & \textbf{74.5} & 73.0 & 24.9 \\
\Xhline{2\arrayrulewidth} % Thicker line
 & DET & 57.6 & 52.2 & 28.9 & 43.1 & 38.7 & 40.8 \\
 & PROB & 62.6 & 62.4 & 37.1 & 50.3 & 50.7 & 49.1 \\ 
 % \cline{2-14}
 & SAC & 62.7 & 64.9 & 41.5 & 49.7 & 50.3 & 50.8 \\
 & DDPG & 57.7 & 52.9 & 30.0 & 36.2 & 31.4 & 39.9 \\
\parbox[c][0pt][c]{\linewidth}{
  \centering
  \rotatebox[origin=c]{90}{
    \parbox{3cm}{\centering ISMRM}
  }
}
& TD3 & 52.2 & 43.2 & 22.3 & 36.9 & 31.6 & 39.0 \\
 & TRLF$_{S}$ & 61.1 & 58.2 & 32.1 & 49.1 & 51.1 & 57.3 \\
& $\pi_{maxQ}$ & 62.8 & 65.1 & 41.6 & 50.2 & 49.8 & 48.8 \\
& $\pi_{avg}$ & 62.8 & 59.5 & 30.1 & 46.7 & 45.4 & 49.1 \\
 %mod----
 & \textbf{FusionNet} & \textbf{63.9} & 65.4 & 39.1 & \textbf{52.6} & 54.3 & 51.2 \\ \cline{1-8}
 \hline
\end{tabular}
% }
% \end{center}
% \caption{Commands that must not be used}

\renewcommand{\arraystretch}{1.0}
% ============ (b) ISMRM (Original Style) ============
\textbf{(b) AF and CC}

\begin{tabular}{ |p{0.44cm}|p{1.3cm}|p{0.5cm}p{0.5cm}p{0.5cm}|p{0.5cm}p{0.5cm}p{0.5cm}| }

 \hline
  &  & \multicolumn{3}{c|}{\textbf{AF}} & \multicolumn{3}{c|}{\textbf{CC}} \\ \cline{3-8}
\textbf{Data} & \textbf{Algorithm} & \textbf{\textbf{Dice}\textuparrow} & \textbf{OL}\textuparrow & \textbf{OR}\textdownarrow & \textbf{Dice}\textuparrow & \textbf{OL}\textuparrow & \textbf{OR}\textdownarrow \\ \cline{1-8}
 % & TractSeg &  &  &  &  &  &  &  &  &  &  &  &  \\
 & DET & 53.4  & 41.3 & 10.3 & 70.9 & 77.2 & 41.2 \\
 % & PROB & \tikzmark{CG hcp 1 top}56.9 & 48.7 & 22.3\tikzmark{CG hcp 1 bottom} & \tikzmark{CST hcp 1 top}74.7 & 71.8 & 20.0\tikzmark{CST hcp 1 bottom} & \tikzmark{AF hcp 1 top}56.5 & 45.8 & 13.5\tikzmark{AF hcp 1 bottom} & \tikzmark{CC hcp 1 top}74.9 & 84.1 & 41.6\tikzmark{CC hcp 1 bottom} \\ \cline{2-14}
 & SAC & 54.6 & 42.4 & 11.4 & 70.4 & 80.4 & 48.1 \\
 & DDPG & 50.6 & 37.9 & 8.3 & 65.0 & 66.2 & 37.3 \\
% \parbox[c][0pt][c]{\linewidth}{\centering\rotatebox[origin=c]{90}{HCP \shortcite{wasserthal2018tractseg}}}

\parbox[c][0pt][c]{\linewidth}{
  \centering
  \rotatebox[origin=c]{90}{
    \parbox{3cm}{\centering HCP}
  }
} & TD3 & 51.7 & 39.5 & 9.6 & 64.9 & 66.2 & 37.2 \\
 & TRLF$_{S}$ & 50.9 & 38.0 & 8.2 & 67.6 & 67.5 & 30.9 \\
  & $\pi_{maxQ}$ & 54.8 & 42.3 & 10.9 & 70.4 & 81.6 & 50.1 \\
  & $\pi_{avg}$ & 52.3 & 40.5 & 12.7 & 70.1 & 78.4 & 46.1 \\
 %mod---- yoooo
 & \textbf{FusionNet} & \textbf{55.5} & 43.8 & 12.3 & \textbf{72.4} & 81.8 & 44.8 \\ \cline{2-8}
 & TractSeg & 70.9 & 59.5 & 8.1 & 76.8 & 68.5 & 11.7 \\
 & \textbf{FusionNet*} & \textbf{74.1} & 66.6 & 13.0 & \textbf{77.4} & 78.6 & 23.9 \\
  % & SAC & 56.2 & 47.3 & 20.2 & 67.7 & 60.8 & 17.4 & 54.6 & 42.4 & 11.4 & 70.4 & 80.4 & 48.1 \\
\Xhline{2\arrayrulewidth} % Thicker line
 & DET & 50.0 & 44.0 & 35.8 & 49.7 & 39.5 & 19.6 \\
 % & PROB & \tikzmark{CG ttoi 1 top}66.3 & 66.9 & 35.2\tikzmark{CG ttoi 1 bottom} & \tikzmark{CST ttoi 1 top}76.2 & 76.0 & 23.3\tikzmark{CST ttoi 1 bottom} & \tikzmark{AF ttoi 1 top}56.3 & 52.8 & 39.1\tikzmark{AF ttoi 1 bottom} & \tikzmark{CC ttoi 1 top}59.7 & 53.2 & 25.9\tikzmark{CC ttoi 1 bottom} \\ \cline{2-14}
 & SAC & 52.3 & 52.4 & 52.1 & 54.6 & 48.3 & 28.6 \\
 & DDPG & 46.3 & 39.3 & 28.6 & 49.6 & 40.4 & 22.1 \\
\parbox[c][0pt][c]{\linewidth}{
  \centering
  \rotatebox[origin=c]{90}{
    \parbox{3cm}{\centering TractoInferno}
  }
} & TD3 & 45.3 & 39.6 & 37.6 & 48.1 & 38.3 & 20.9 \\
& TRLF$_{S}$ & 47.5 & 43.5 & 43.2 & 46.9 & 37.0 & 21.1 \\
 & $\pi_{maxQ}$ & 52.7 & 51.9 & 47.3 & 54.2 & 47.0 & 30.7 \\
  & $\pi_{avg}$ & 52.1 & 51.9 & 49.5 & 53.3 & 47.3 & 31.5 \\
 %mod----
 & \textbf{FusionNet} & \textbf{53.2} & 51.3 & 39.1 & \textbf{56.5} & 50.5 & 28.8 \\
 \hline
\end{tabular}
\vspace{-0.2cm}
\caption{Performance comparison (\%) for CG, CST/PYT, AF, CC tracts on HCP, TractoInferno, and ISMRM datasets. Dice, OL, OR are averaged across test subjects. DET, PROB results from \cite{joshi2024tract}; SAC, DDPG, TD3 from \cite{theberge2024matters} were used to train our network, compared with same specifications. SAC based TRLF$_S$\cite{joshi2024tract} trained on EDS. TractSeg and FusionNet* use TractSeg masks, evaluated on HCP. Best Dice is in bold.}
% Added by Ashutosh-end
% Average Scores are reported across all test subjects. 
\label{table:exp:combined}
% \vspace{-0.3cm}
% \vspace{-2cm}
\end{table}
% \vspace{-0.3cm}
This section presents evaluation on test subjects from TractoInferno (1160, 1078, 1061, 1159, 1171), HCP (930449, 959574, 992774, 987983) and ISMRM.
We compare with: Classical DET \cite{basser2000vivo} and PROB (iFOD1) \cite{tournier2012mrtrix}; RL policies SAC, TD3, and DDPG\cite{theberge2021track,theberge2024matters} employed in TractRLFusion, and their decision-level ensembles $\pi_{maxQ}$ (Max Q-value action) and $\pi_{Avg}$(average of actions),
TractSeg \cite{wasserthal2018tractseg} and \cite{joshi2024tract}, incorporating SAC (best performing) in their framework (TRLF$_{S}$). 
Since PROB contributed to create TractoInferno ground truth, it may be inherently favored. Moreover, reference streamlines for HCP \cite{wasserthal2018tractseg} are made using iFOD2, a close successor of PROB. Hence it is evaluated on ISMRM.
TractOracle \cite{levesque2025exploring} was left out, as it is tailored for WM/GM–interface seeding. Comparative analysis reveal the following notable insights:

% can be influenced by the choice of segmentation tool.
% TractOracle \cite{theberge2024tractoracle} was excluded from our comparisons, as it is specifically designed for seeding from the white matter–gray matter (WM/GM) interface.

% Comparative results in Table \ref{table:exp:combined} and qualitative results in Fig. \ref{fig:plot} reveal the following notable insights:

\begin{figure}[H]
    \centering
    \includegraphics[width=7.5cm]{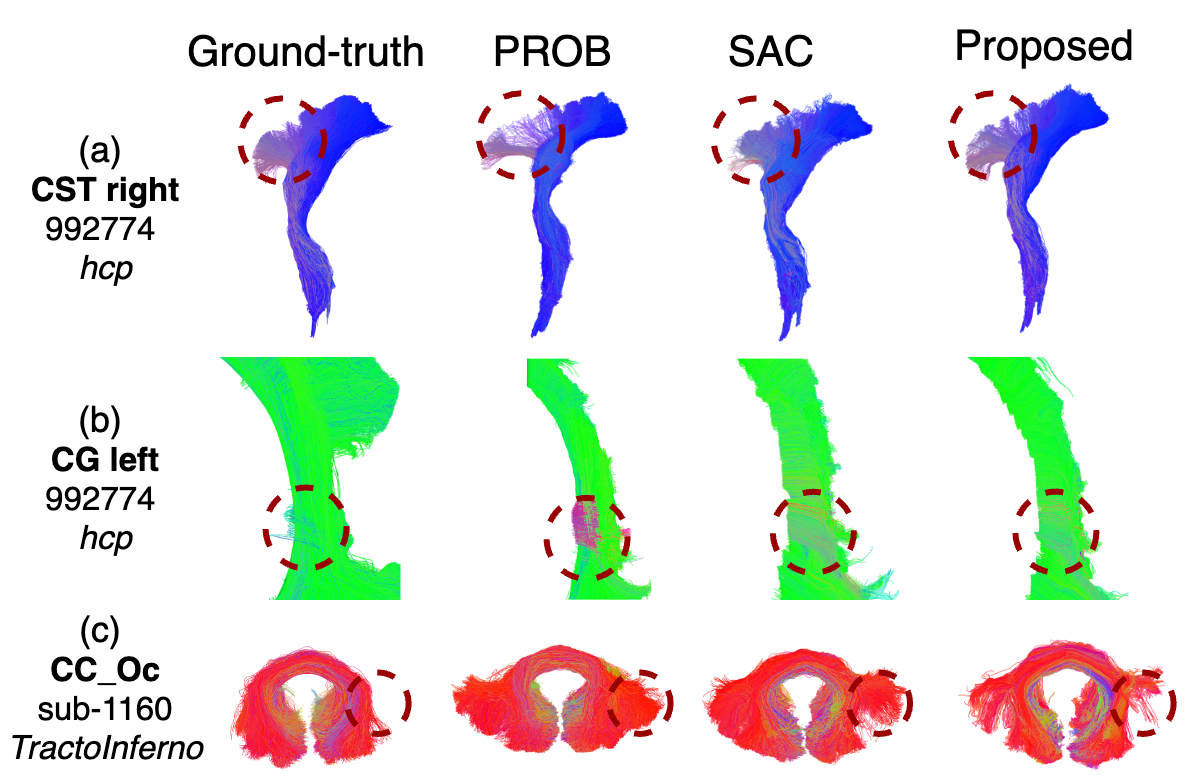}
    % \caption{Visualization of (a) right corticospinal tract, and (b) part of the left cingulum tract, from subject 992774 in the HCP dataset. (c) shows the occipital part of the corpus callosum from subject 1160 in the TractoInferno dataset; for comparison of proposed method with PROB and SAC (other top-performing algorithms considered in this study) and ground-truth tracts. Dashed circles highlight key regions for comparison.}
    % \caption{Visualization of (a) right corticospinal tract, and (b) part of the left cingulum tract, from subject 992774 in the HCP dataset. (c) shows the occipital part of the corpus callosum from subject 1160 in the TractoInferno dataset; Compares proposed method with PROB and SAC (other top-performing algorithms considered here) and ground-truth tracts. Dashed circles highlight key regions for comparison.}
    % \caption{Visualization of (a) right CST tract, and (b) part of the left CG, from subject 992774 in the HCP dataset. (c) shows the occipital part of CC from subject 1160 in the TractoInferno dataset; Compares proposed method with PROB and SAC (other top-performing algorithms) and ground-truth tracts. Dashed circles highlight key regions for comparison.}
    
    \caption{Visualization of the right CST (a) and a segment of the left CG (b) from HCP subject 992774, and the occipital CC (c) from TractoInferno subject 1160. The proposed method is compared with PROB, SAC (other top-performers), and ground-truth tracts, with dashed circles marking key regions.}
    \label{fig:plot}
    \vspace{-0.5cm}
\end{figure}

% \textbf{Policy Ensembles} $\pi_{maxQ}$ and $\pi_{Avg}$ of SAC, TD3, and DDPG at decision level by selecting the action with the max normalized Q-value, and average of the 3 actions, respectively, at each step.
\begin{itemize}%[label=--, leftmargin=*, itemsep=0.2em]
% above label, leftmargin, itemsep is commented by Ashutosh
    % \item FusionNet (Proposed) consistently ranks either first or second alongside PROB, showcasing best performance in ISMRM dataset CG and CST [Table \ref{table:exp:combined}].
    \item Our FusionNet model consistently ranks best across all datasets and tracts  as shown in Table \ref{table:exp:combined}.
    % \item \textit{Note:} Since PROB contributed to create TractoInferno ground truth, it may be inherently favored. Reference streamlines for HCP \cite{wasserthal2018tractseg} are made using iFOD2, a close successor of PROB. Hence it is evaluated on ISMRM.
% Therefore its performance on ISMRM is presented.
    % and thus it may systematically underperform on it. Its performance on ISMRM is presented.
    % \item Fusion policy (FusionNet) outperforms individual RL policies (SAC, TD3, DDPG) which show a high OR-OL tradeoff with SAC achieving highest OL with worst OR, and DDPG and TD3 achieving low OL and OR. FusionNet balances the OL-OR trade-offs resulting in the \textbf{best Dice}.
    % Added by Ashutosh-begin
    % \item Fusion policy (FusionNet) outperforms individual RL policies (SAC, TD3, DDPG) and the Ensemble method, which combines SAC, TD3, and DDPG via highest Q-value selection. Individual RL policies show a high OR-OL tradeoff with SAC achieving highest OL with worst OR, and DDPG and TD3 achieving low OL and OR. The Ensemble’s Q-value approach yields suboptimal tracking, while FusionNet’s learned policy better balances OL-OR trade-offs, resulting in the \textbf{best Dice}.
    \item Fusion policy (FusionNet) outperforms individual RL policies (SAC, TD3, DDPG) and their RL-Ensemble methods ($\pi_{maxQ}$ and $\pi_{avg}$). Individual RL policies show a high OR-OL tradeoff with SAC achieving highest OL with worst OR, and DDPG and TD3 achieving low OL and OR. The Ensemble’s Q-value approach yields suboptimal tracking, while FusionNet’s learned policy better balances OL-OR trade-offs, resulting in the \textbf{best Dice}.
    \item FusionNet* (FusionNet tested on TractSeg masks \cite{wasserthal2018tractseg}) surpasses TractSeg across all HCP tracts, e.g., achieving a Dice score of 74.1 versus TractSeg’s 70.9 for AF, demonstrating superior tracking even within TractSeg’s masks.
    \item FusionNet* and TractSeg achieve highest scores on HCP, as TractSeg was trained on HCP data to obtain those masks which were also involved in making reference streamlines of HCP. However, masks from~\cite{joshi2024tract} were available for all three datasets (TractoInferno, HCP, and ISMRM), preferred for a consistent comparison across these datasets. FusionNet’s superior performance within both masks demonstrates its robustness to variations in mask quality for tract-specific tractography.
    
    % as its model was pretrained on HCP data, producing masks aligned with reference tracts. 
    % Added by Ashutosh-end
    \item Spurious connections made by SAC and PROB in left CG and CC\_Oc encircled in red (Fig. \ref{fig:plot}(a,b)) show lower OR by our method, consistent with the scores in table \ref{table:exp:combined}.
    \item In Fig. \ref{fig:plot}(a), PROB struggles to reconstruct upper portion of CST right, at pyramidal decussation (intersection of left \& right CST) compared to FusionNet.
    \item Similar visual trend was observed in these tracts for other subjects, but we limit for brevity. 
\end{itemize}
% \vspace{-0.2cm}
\textbf{\textit{Note:}} While simpler ensembling methods ($\pi_{maxQ}$ and $\pi_{Avg}$) achieve lower scores than FusionNet, more complex RL ensembling techniques often introduce stability and scalability challenges. TractRLFusion addresses (this) fusion by learning a robust fusion policy in a data-driven manner, which simplifies integration compared to more traditional RL ensemble methods. 
We illustrate our method with three policies (SAC, DDPG, TD3) to address the overlap–overreach trade-off, but the framework is modular and can incorporate additional policies to capture broader tracking behaviors.

% Moreover, while we demonstrate our method using three policies (SAC, DDPG, and TD3) to handle the overlap-overreach trade-off (observed in these individual policies), the proposed framework is modular and scalable. It can be extended to include additional policies to further capture diverse tracking behaviors.
% and improve generalization.

% \vspace{-2cm}

\textbf{Ablation Study}: The ablation study underlines the importance of EDS and MCPFT. EDS involves a within-policy and across-policy (EDS$_{across-\pi}$) selection of training trajectories, whereas EDS$_{across-\pi}$ selects trajectories across different policies using only Q-value. Table \ref{table:abl:combined} reveals that pretraining and finetuning on only EDS (\textit{ii}) performs worse than (EDS$_{across-\pi}$)  (\textit{i}). Pretraining on coarser EDS$_{across-\pi}$ and finetuning on EDS (\textit{iii}) performs the best. Hence FusionNet is trained on data from (iii), followed by MCPFT, whose results in Table \ref{table:exp:combined} show effectively lower OR with higher OL, a desirable outcome in tractography, leading to a higher Dice.

\begin{table}[H]
% [hbt!]
\begin{center}
% \caption{Average performance (\%) obtained using Fusion-Net on test subjects from the TtoI. The table compares different pretraining (PT) and finetuning (FT) strategies and highlights the impact of incorporating tract-specific information using EDS and Multi-Critic Policy Fine-Tuning (MCPFT).}
% \fontsize{8}{7.5}\selectfont
\fontsize{9}{10}\selectfont
% \centering
\vspace{-0.2cm}
\begin{tabular}{|>{\centering\arraybackslash}m{0.54cm}|p{0.4cm}p{0.4cm}p{0.4cm}|p{0.4cm}p{0.4cm}p{0.4cm}|p{0.4cm}p{0.4cm}p{0.4cm}|}
\hline
\textbf{Tract}&\multicolumn{3}{c|}{{\makecell{\textbf{(i) EDS$_{\text{across}-\pi}$} \\ (PT \& FT)}}} & \multicolumn{3}{c|}{{\makecell{\textbf{(ii) EDS} \\ (PT \& FT)}}} & \multicolumn{3}{c|}{\makecell{\textbf{(iii) EDS$_{\text{across}-\pi}$} \\  PT \& EDS FT}}\\
\cline{2-10}
& Dice& OL& OR& Dice & OL& OR& Dice& OL & OR\\
\hline
PYT & 70.4 & 67.3 & 23.4 & 58.6 & 45.7 & 8.9 & 70.9 & 68.3 & 23.9 \\
CG  & 59.2 & 58.6 & 39.4 & 42.9 & 32.1 & 10.7 & 61.4 & 59.5 & 34.5 \\
AF  & 47.3 & 39.7 & 27.3 & 23.3 & 14.4 & 5.3 & 48.7 & 43.5 & 38.9 \\
CC  & 67.7 & 74.1 & 44.2 & 60.6 & 50.3 & 14.9 & 68.6 & 70.2 & 36.4 \\
\hline
\end{tabular}
\end{center}
\vspace{-0.5cm}
\caption{Mean (\%) FusionNet scores on TractoInferno test data comparing pretraining (PT) \& finetuning (FT) strategies.}
% , showing the effect of EDS.}
% \caption{Mean (\%) scores of FusionNet on TractoInferno test subjects, comparing different pretraining (PT) and finetuning (FT) strategies, highlighting the impact of EDS.}
% and Multi-Critic Policy Fine-Tuning (MCPFT).}
\label{table:abl:combined}
\vspace{-0.3cm}
\end{table}

\vspace{-0.6cm}
\section{Conclusion}
\vspace{-0.2cm}
\textbf{TractRLFusion} enables efficient and effective white matter tracking by fusing complementary strengths of different actor-critic frameworks. Our ablation studies highlight the effectiveness of the proposed two-stage data curation, where a strong baseline policy is first developed using a combination of broad and anatomically-aware data selection (EDS), and subsequently refined through the MCPFT process. TractRLFusion demonstrates strong \textbf{generalization}, being trained solely on TractoInferno and evaluated on HCP and ISMRM.
Finally, powered by \textbf{FusionNet},  \textbf{TractRLFusion} lays the groundwork for a foundational tractography model with the potential to extend across additional datasets and integrate diverse specialized policies, advancing the field of White Matter Tractography.

\vspace{-0.3cm}
\section{Compliance with ethical standards}
This study used open-access human data from TractoInferno \cite{poulin2022tractoinferno} and the Human Connectome Project \cite{wasserthal2018tractseg} in accordance with their licenses, and publicly available phantom data from the ISMRM 2015 Tractography Challenge \cite{maier2017challenge}.
% , which involves no human or animal subjects.
% This work has been conducted using human subject data made available in open access by TractoInferno \cite{poulin2022tractoinferno}, Human Connectome Project \cite{wasserthal2018tractseg}, as confirmed by the license attached with the open access data. Additionally, this study used phantom diffusion MRI data from the ISMRM 2015 Tractography Challenge \cite{maier2017challenge}, which does not involve human or animal subjects and is publicly available for research use.

% \section{Acknowledgments}
% \label{sec:acknowledgments}

\section{Acknowledgments}
\label{sec:acknowledgments}
This work was supported by SERB Core Research Grant Project No: CRG/2020/005492, IIT Mandi.
% This work was supported by IIT Mandi from SERB CORE Research Grant with Project No: CRG/2020/005492.

\bibliographystyle{IEEEbib}
\setlength{\bibsep}{0pt plus 0.3ex}
\renewcommand{\bibfont}{\small}
\vspace{-0.4cm}
\bibliography{strings,refs_squeezed}
%\bibliography{strings,refs}

\end{document}